\documentclass[conference,letterpaper,twoside,final,10pt]{ieeeconf}
\usepackage[left=1.91cm,right=1.91cm,top=1.91cm,bottom=1.91cm]{geometry}

\usepackage{graphicx}
\usepackage{amssymb}
\usepackage{amsmath}
\usepackage{cite}
\usepackage{comment}
\usepackage{color}
\usepackage{url}
\usepackage{alltt}
\usepackage{cleveref}
\usepackage{tikz}
\usepackage{pgfplots}
\usepackage{pgf-umlsd}
\usepackage{cprotect}
\usepackage{multirow}
\usepackage{algorithm}
\usepackage[noend]{algpseudocode}
\usepackage{bigfoot}
\usepackage{paralist}
\usepackage{tabularx}
\usepackage[whole]{bxcjkjatype}
\usepackage{ifthen}
\usepackage{threeparttable}
\usepackage[hang,small,bf]{caption}
\usepackage[subrefformat=parens]{subcaption}
\usepackage[super]{nth}
\usepackage{colortbl}
\usepackage{scalefnt}
\usepackage{threeparttable}
\usepackage{multicol}
\usepackage{booktabs}

\pgfplotsset{compat=1.14}
\IEEEoverridecommandlockouts

\makeatletter
\newcommand{\figcaption}[1]{\def\@captype{figure}\caption{#1}}
\newcommand{\tblcaption}[1]{\def\@captype{table}\caption{#1}}
\makeatother

\newcommand{\red}[1]{\textcolor{black}{#1}} 

\title{\LARGE \bf
Cluttered Food Grasping with Adaptive Fingers \\and Synthetic-Data Trained Object Detection
}

\author{
Avinash Ummadisingu$^{\dagger}$,
Kuniyuki Takahashi$^{\dagger}$,
Naoki Fukaya$^{\dagger}$
\thanks{$^{\dagger}$ A. Ummadisingu, K. Takahashi, and N. Fukaya are associated with Preferred Networks, Inc.
 \ \{ummavi, takahashi, fukaya\}@preferred.jp}}

\begin{document}

\maketitle
\thispagestyle{empty}

\begin{abstract}
The food packaging industry handles an immense variety of food products with wide-ranging shapes and sizes, even within one kind of food. Menus are also diverse and change frequently, making automation of pick-and-place difficult.
A popular approach to bin-picking is to first identify each piece of food in the tray by using an instance segmentation method.
However, human annotations to train these methods are unreliable and error-prone since foods are packed close together with unclear boundaries and visual similarity making separation of pieces difficult.
To address this problem, we propose a method that trains purely on synthetic data and successfully transfers to the real world using sim2real methods by creating datasets of filled food trays using high-quality 3d models of real pieces of food for the training instance segmentation models.
Another concern is that foods are easily damaged during grasping.
We address this by introducing two additional methods- a novel adaptive finger mechanism to passively retract when a collision occurs, and a method to filter grasps that are likely to cause damage to neighbouring pieces of food during a grasp.
We demonstrate the effectiveness of the proposed method on several kinds of real foods.
\end{abstract}
\section{Introduction}
\label{sec:introduction}
Convenience store meals are ubiquitous and continue to see fast-growing popularity globally. Despite the growing need for automation in the food-packing industry, its adoption sees limited success due to numerous factors. A major impediment to automation is the remarkable diversity in the types of foods handled and significant variance in shapes and sizes, even between pieces of the same food. To further complicate the issue, menus frequently change due to seasonality or customer interest. Custom-designed hardware and software for handling individual foodstuff see limited applicability due to the need to adapt to novel foods quickly. 

A popular approach to bin-picking is using object detection algorithms to identify the objects in the bin, derive a grasp pose, and then plan and execute it~\cite{kleeberger2020survey}. However, collecting large datasets required to train these object detection methods for food is challenging due to the high costs of collecting data and annotations, as well as dealing with mistakes made due to the high visual similarity and difficulty in obtaining clear boundaries between pieces packed together in a bin. In addition, foods are fragile and easily damaged by interaction and change in behavior over time due to environmental factors (temperature, humidity) as well as loss of water content and spoilage.
It makes the gathering of large datasets of real food challenging and often requires techniques to learn from limited samples~\cite{takahashi2021uncertainty, takahashi2021target}.

\begin{figure}[t]
	\centering
	\includegraphics[width=0.8\columnwidth]{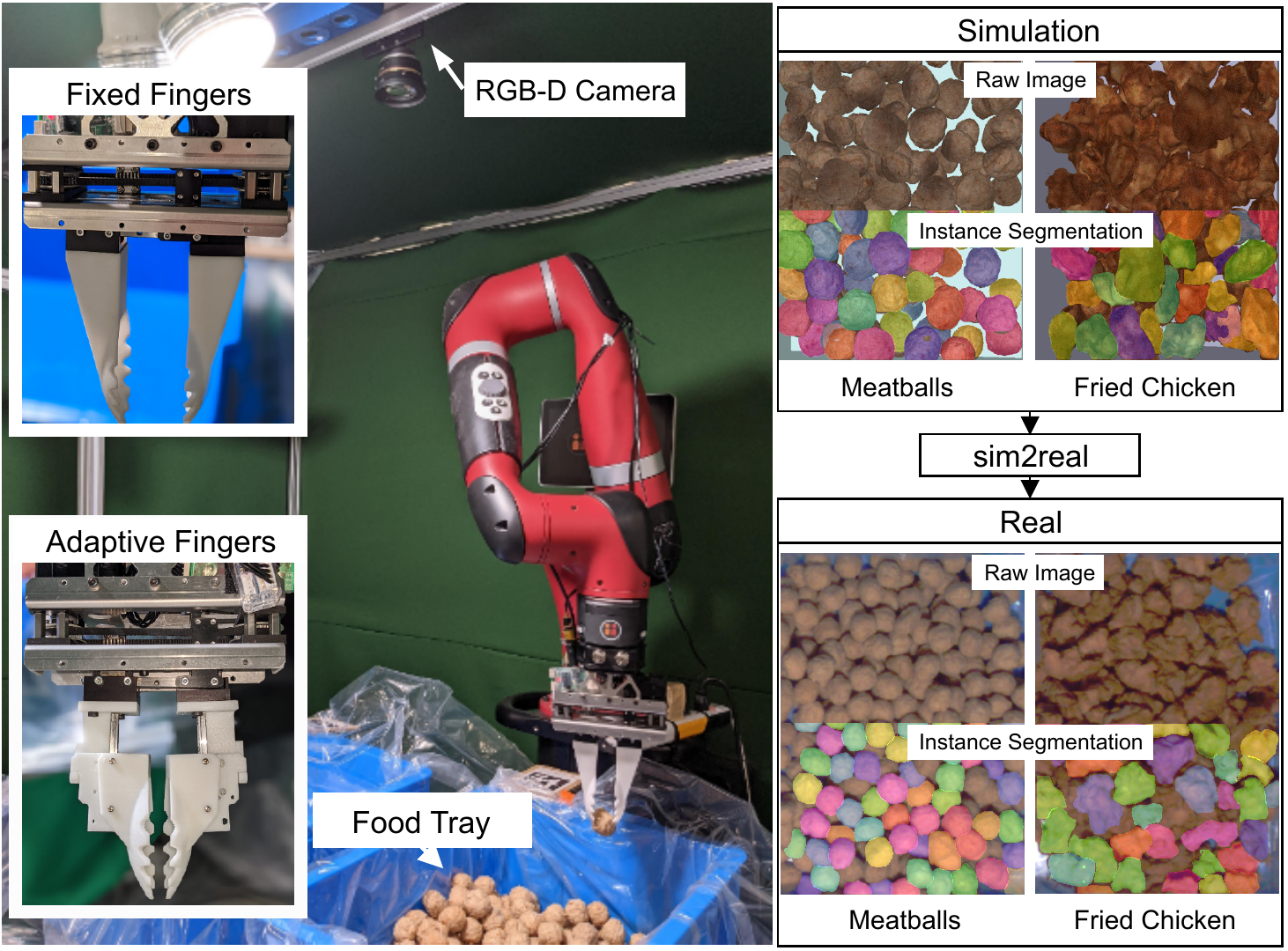}
	\caption{Experiment Setup: Robot setup for the experiments with instance segmentation detections trained on synthetic images and used for grasping in the real-world.}
	\label{fig:setup}
\end{figure}

To address this problem, we propose \textbf{1)} using synthetic data generated in simulation using highly realistic 3d models of real food to train an instance segmentation model. By applying ideas of sim2real and domain randomization~\cite{tremblay2018training}, we enable the model trained on synthetic data to transfer to the real world without additional training, overcoming the cost and problems faced in collecting human annotations (Fig.~\ref{fig:setup}).

Another issue that sets apart the grasping of food from existing methods is their fragility. 
Foods may be soft, delicate and easily damaged if grasped with excessive force incurring an unacceptable cost. This is especially likely since the grasping is done from trays packed full of food. In this work, we propose two methods to alleviate this- \textbf{2)} The introduction of a novel adaptive finger mechanism that passively retracts when it encounters the surface of the food, and \textbf{3)} a grasp filtering heuristic that filters risky grasp candidates that are likely to collide with neighbouring foods and damage them.


\begin{figure*}[t]
	\centering
	\includegraphics[width=1.4\columnwidth]{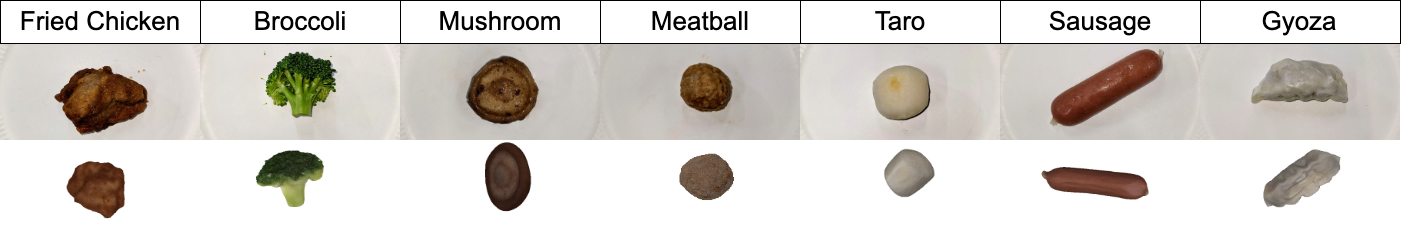}
	\caption{Real foods used for experiments and their corresponding 3d models the segmentation model was trained with.}
	\vspace{-7mm}
	\label{fig:simulation_3dmodel}
\end{figure*}
\section{Related Work}
\label{sec:related works}
\subsection{Instance Segmentation of food}
\label{sec:instancesegmentationfood}
There exist several labelled datasets for foods, often under the field of food computing~\cite{min2019survey}.
Existing food image segmentation datasets, such as Foood201-Segmented~\cite{meyers2015im2calories}, UECFoodPix~\cite{uecfoodpix}, and others~\cite{uecfoodpixcomplete, ciocca2016food, gao2019musefood, aslan2020benchmarking, park2021deep, wu2021large} provide instance-level labels but are not at the granularity required for our application (a plate full of pieces of chicken would be identified as one instance).

The gathering of datasets of the size required for deep learning models are unavoidably labor-intensive tasks.
Prior works have explored the use of simulation or synthetic data for training semantic or instance segmentation models in numerous application such as crop seed detection~\cite{toda2020training} and even food~\cite{park2021deep}. However, they use simple shapes and focus on detecting food areas like the datasets above.
Advances in photogrammetry and techniques such as differentiable rendering~\cite{kato2020differentiable} allow for easy capture and generation of near photo-realistic models of objects from a handful of images, eliminating the need for specialist knowledge to create them. 
In our work, we use these techniques for quick and easy capture of novel foods that can be used in simulation to generate synthetic data to identify individual pieces of food.

\subsection{Food Grasping from Hardware Perspective}
\label{sec:learninggrasp}
Rapid R\&D and adoption of soft grippers that reduce damage to objects or its surroundings is ongoing.
The gripper fingers are soft and do not require precise control, but adjust with the food passively~\cite{dimeas2015robotic, endo2020robotic, gafer2020quad, dang2021robotic}.
However, the fingers tend to be thick, which makes it difficult to grasp food that is tightly packed in a tray.
Grasping by suction can be done even in a cramped setting, but is challenging to use due to concerns of hygiene and contamination of the vacuum hose~\cite{sam2010robotic, morales2014robotic, elango2018robotic}.
Prior work has also looked at grasping a single object with a needle gripper, but the needle is inserted into the food, raising two concerns- damage to the food and hygiene~\cite{wang2021soft}. 
In our work, we develop a parallel gripper with an adaptive finger mechanism that passively retracts when it encounters the surface of the food to prevent damage.
\subsection{Collision Avoidance in Grasping}
\label{sec:Collision Avoidance in Grasping}
Grasping in clutter makes the task of collision avoidance significantly harder. This is of particular importance to food since collisions could cause undue damage to foodstuff.
There are three main approaches: 1) heuristic-based, 2) sampling-based, and 3) data-driven.
1) Heuristic-based approaches such as finding the highest grasp points~\cite{wang1994model} are simple but powerful if the appropriate one is selected for the task.
2) Sampling-based approaches evaluate grasp candidates for collision by using analytical methods~\cite{spenrath2017gripping, fan2019efficient, mano2019fast,kleeberger2020transferring, xue2008automatic} or by using task-and-motion planning libraries like OpenRAVE for collision checking~\cite{diankov2010automated, ten2017grasp, dogar2010push, kitaev2015physics, berenson2007grasp, ciocarlie2014towards}. These approaches are computationally expensive to do at the resolution needed to grasp small pieces of food.
3) Data-driven approaches are popular in robotics with deep learning~\cite{fang2020graspnet, kleeberger2019large}.
These methods require a large amount of data or learn in simulation~\cite{ten2017grasp, mahler2017learning,wang2021hierarchical,lundell2021ddgc, li2021simultaneous,lou2021collision,  murali20206}. However, collection of a large manipulation dataset of food is not possible and realistic simulation of soft-bodied objects like food is still challenging to apply.

Our approach belongs to 1) heuristic-based.
The method closest to our own is that of \cite{domae2014fast} where they represent a gripper model by two mask images representing a contact region and a collision region (which should be free of objects). However, they focus on rigid objects with regular shapes and are able to enforce collision regions to be completely empty. However, in dealing with soft, irregular foodstuffs, collisions to gently push food are unavoidable for a successful grasp.

\begin{figure*}[t]
	\centering
	\includegraphics[width=1.80\columnwidth]{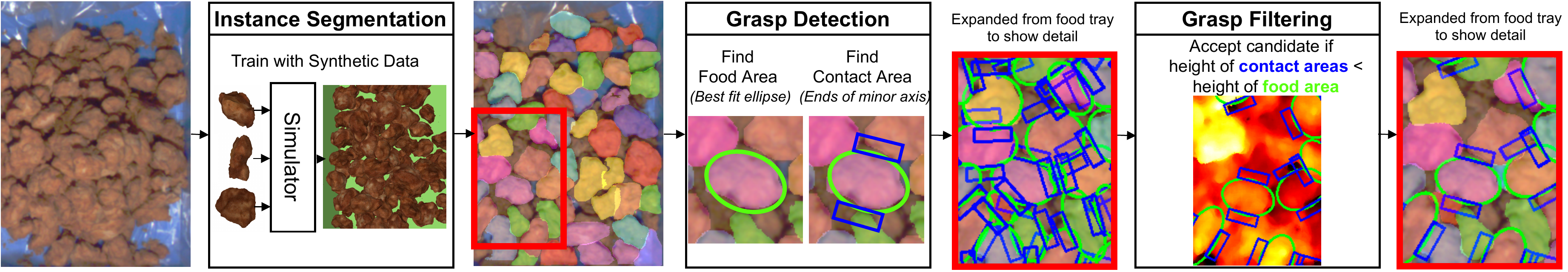}
	\caption{Grasping method that avoids damaging food to be grasped and surrounding pieces. The method is composed of instance segmentation, grasp detection, and grasp filtering. The green circles represent the best-fit elliptical mask for each identified instance. The blue rectangles represent candidate contact points between the gripper's fingers and the food, and are placed around the ends of the minor axis of the ellipse.}
	\label{fig:pipeline}
	\vspace{-5mm}
\end{figure*}

\begin{figure}[t]
	\centering
	\includegraphics[width=0.8\columnwidth]{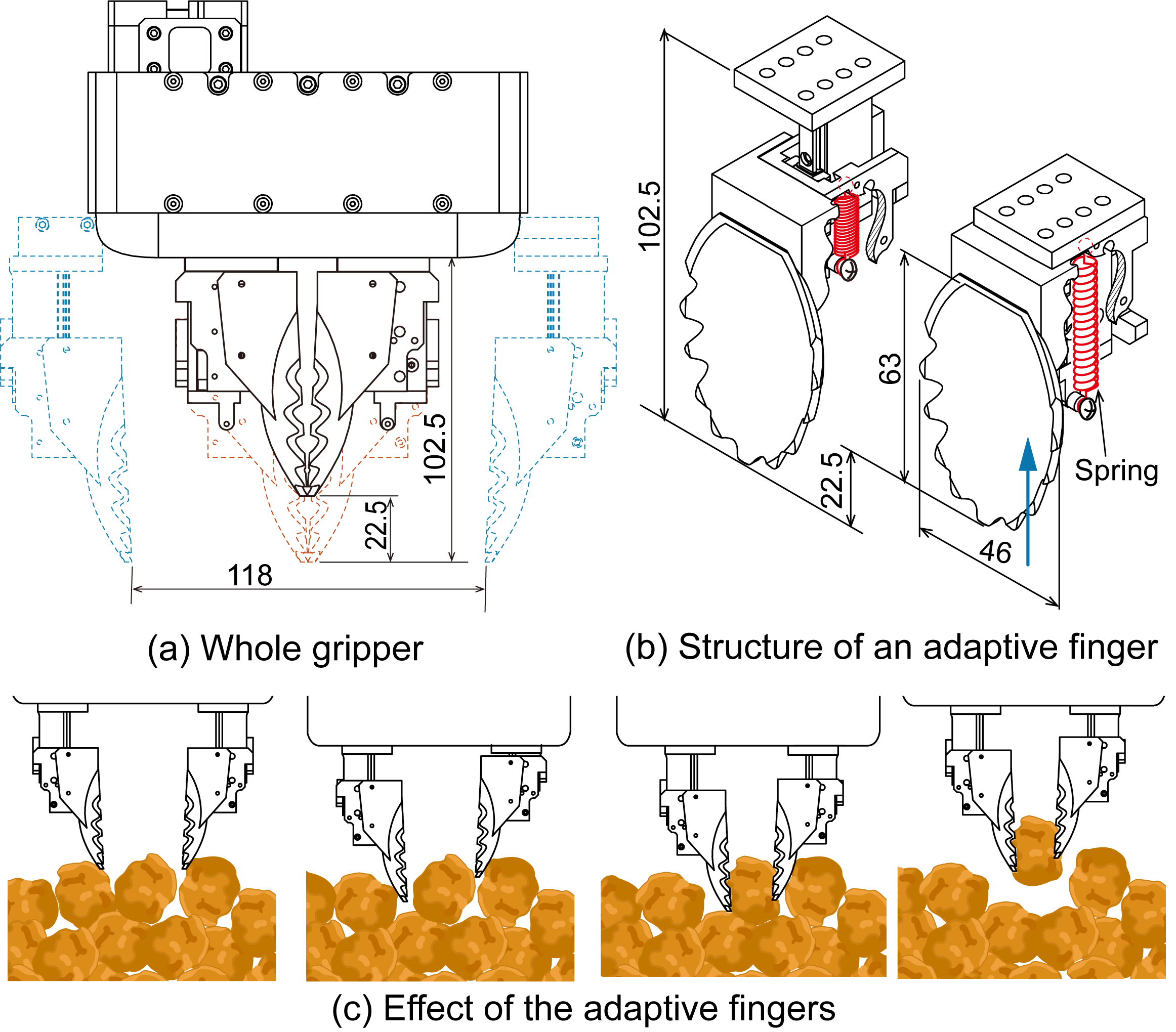}
	\caption{Schematic diagram of the adaptive finger with a passive retraction mechanism to avoid damaging food on insertion.}
	\label{fig:soft_schematic}
\end{figure}
\section{Method}
\label{sec:method}
Our proposed method to grasp a piece of the target food from a cluttered tray consists of the following components:
A) Generate \emph{synthetic images} and train with them, B) \emph{grasp filtering} for collision avoidance during grasping, and C)  grasp using \emph{adaptive fingers}.
\subsection{Synthetic Image Generation \& Training}
\label{sec:synimage}
In this section, we describe how we create a training dataset to train an instance segmentation model capable of detecting individual pieces of food in a cluttered tray.
We first create a high-quality 3d model from several photographs of each piece of food using Preferred Networks, Inc.'s technology~\cite{pfn2021}.
The details of the 3d model generation are outside the scope of this work.
Fig.~\ref{fig:simulation_3dmodel} shows the real foods and their corresponding 3d models.
Highly uniform foods, such as sausage, can easily be represented by a single model, while we prepare the capture of multiple pieces for foods that show variance in shapes and sizes, such as fried chicken. 
We model a tray in simulation of the same size as the one shown in Fig.~\ref{fig:setup}, and drop a number of these captured 3D models into the tray to generate training data for instance segmentation.
We vary the number of pieces generated and initialize them at random locations over the tray.
In doing this, we depend on the physics engine to prevent the models from conjoining and resolving collisions. Detailed parameters are described in Section~\ref{sec:Synthetic data generation}

Although we can capture high-quality models of the foods and tune the simulation to be as realistic as possible, photorealism to the level needed to transfer to the real world is unattainable.
To improve the model's generalization, we adopt concepts of domain randomization from sim2real literature to randomly vary simulation parameters: the tray color, scale of the model, lighting parameters (direction, ambient light, shadows), and direction of the camera. 
Using domain randomization allows the model to traverse the reality gap and be used directly for real-world predictions without the need to additionally fine-tune with actual food data.
\subsection{Grasp Detection \& Filtering}
\label{sec:grasp_detection}
In this section, we explain the method used to find a grasp point and pose for grasping a piece of food while avoiding damage to it or its neighbours when the gripper is inserted.
The method consists of three steps; 1) Instance segmentation, 2) grasp detection, and 3) grasp filtering (Fig.~\ref{fig:pipeline}).

\subsubsection{Instance Segmentation}
We use an instance segmentation model trained using images generated as per the previous section. We pass an RGB image of the food tray as input to the model and obtain a mask for each piece of food.  \red{The use of depth for segmentation was omitted due to difficulty in adequately matching generated depth to reality and the lack of improved performance at the cost of inference speed.}

\subsubsection{Grasp Detection}
We employ a method of ellipse fitting to segmentation masks to identify a grasp.
The choice to use a best-fit ellipse to represent the food area instead of a typical minimum area rectangle, was motivated by the irregularity in the shapes of the pieces of food. 
For each detected mask (which should, ideally, perfectly correspond to a piece of food), we identify its best fit ellipse (not bounding ellipse). 
We use the parameterization of this best fit ellipse, defined by the 5-tuple $e=\{\mathcal{X}, \mathcal{Y}, \Theta, \mathcal{A}, \mathcal{B}\}$, to derive the grasp parameters of the parallel gripper where $(\mathcal{X}, \mathcal{Y})$ is the center of the ellipse, $\Theta$ its angle of rotation and $(\mathcal{A},\mathcal{B})$ representing the length of its minor and major axis respectively.
We adopt the grasp parametrization of a parallel gripper with the 5-tuple $g=\{x, y, \theta, h, w\}$, which is a common parallel gripper grasp parametrization~\cite{lenz2015deep}.
The grasping point candidate's $(x,y)$ becomes the center of the ellipse $(\mathcal{X}, \mathcal{Y})$.
The width of the minor axis of this ellipse $\mathcal{A}$ becomes the grasp width $w$ and the angle of rotation of the ellipse $\Theta$ gives us the gripper rotation $\theta$. 
The remaining grasp parameter $h$ is calculated from the median height of the \emph{food area} defined by the enclosing best fit ellipse.
These height values are obtained from the RGB-D camera.
To this median height, we add a fixed offset for each food determined experimentally over several trials.

\subsubsection{Grasp Filtering}
From the grasp candidates detected by 2) grasp detection, we apply 3) grasp filtering to filter out candidates that are likely to damage the food piece to be grasped and the surrounding pieces, as well as avoid unintentionally grasping a neighbouring piece, which, depending on the system's application area might be unacceptable.
We represent the \emph{contact area} with the gripper's fingers as two rectangular masks of approximately the same size as the fingers of the gripper.
We then retain only the grasp candidates whose median height of both of these contact areas is lower than the median height of the food area thereby eliminating candidates that cause significant collision with the fingers.
We note that more stringent rules enforcing the entire area be lower or empty would lead to less collision, but some deformable foods may fill up existing free space leading to the filtering step being too harsh in its elimination. In this case, collisions are unavoidable and we need to depend on other means to minimize damage.
From the remaining candidates, we arbitrarily choose the one with the highest median height as the grasp target.
\subsection{Adaptive finger}
\label{sec:Adaptive finger}
This section describes the mechanism of the proposed adaptive finger to grasp fragile foods in a cluttered tray.
Foods may vary in shape and size, and their surfaces are often uneven. This surface irregularity is further exaggerated when a large volume of them is placed in a tray. 
Many foods are fragile and may be easily scratched or damaged during a grasp. Foods such as fried chicken with an outer coating are especially prone to having their surfaces broken.
When grasping such foods with a typical gripper with solid tongs-inspired fingers (which we call \emph{fixed fingers} in Fig.~\ref{fig:setup}), a slight deviation in localization, sensor, or robot control may cause the grasp to fail or the food to be damaged.

In order to address this issue, we have developed a parallel gripper type with an adaptive function to conform to the shape of the food for the fingers (which we call \emph{adaptive fingers} in Fig.~\ref{fig:setup} \& Fig.~\ref{fig:soft_schematic} (a)). This adaptive function allows the gripper's fingers to passively retract when they make contact with the surface of the food thereby avoiding damage caused by the insertion into the food surface.
The fingers are constructed to retract a total of 22.5mm with the spring providing a certain restoring force to the finger.
The contact force of fingertip at maximum shrink is 4.1N (Fig.~\ref{fig:soft_schematic} (b)) \red{and chosen such that is unable to penetrate the surface of foods like fried chicken but still be inserted into grooves between pieces}.
As a result, even if the foods to be grasped are densely packed, the fingertips enter the gap while sliding over the surface of the foods to be grasped, so that it is possible to grasp fragile foods without damaging them (Fig.~\ref{fig:soft_schematic} (c)).
\section{Experiment Setup}
\label{sec:experiments}
\subsection{Model training with Synthetic Data}
\label{sec:Synthetic data generation}
To create an instance segmentation model for grasping each kind of food, we first collect 3d models of the food (Fig.~\ref{fig:simulation_3dmodel}).
We do this by capturing several images and combine them into a high-quality 3d model of the food. 
For some foods like broccoli and fried chicken with remarkable diversity amongst pieces, we capture more than one model in an attempt to improve generalization. 
To further decrease the reality gap, we create a simplified tray modelled after the industrial food tray shown in Fig.~\ref{fig:setup} and position the synthetic camera in a realistic way.
We use the Python bindings of the Bullet physics simulator~\cite{coumans2013bullet,coumans2017pybullet} and employ ideas of domain randomization~\cite{tremblay2018training} to generate images of size $600{\times}600$ to use as training data for the segmentation model.
Pieces of food are dropped from some height over the tray and allowed to fall in order to generate a number of unique, but reasonably realistic images.
We vary a number of simulation parameters to improve generalization, including the number of pieces dropped (10-60), their sizes (0.7x-1.1x), the color of the tray, camera directions and lighting conditions (speculative coefficients, light directions, shadows).

Depending on the complexity of the 3d model, the time required for the physics engine used to generate plausible synthetic data may vary significantly as collision detection and handling becomes more expensive. The average time to generate a single image \red{(using the same machines used to train models specified below)} varies from 10 sec for simple models like sausage, meatball and taro, 14-18 sec for more complex models like fried chicken and mushroom, to 93-459 sec for complex models like gyoza and broccoli.
By breaking generation up into small batches of 10 and deploying them in parallel, we effectively generate the 1200 samples required for each kind of food in around 1 hour.

We train the instance segmentation model using the 1200 samples generated (1000 to train and 200 as the validation set). Additionally, during the training of the instance segmentation model, we apply a number of data augmentations including randomly flipping the images (and their masks) and apply Median Blur to the input images to make the model less reliant on perfectly visible borders.

For the instance segmentation model, we use the highly popular and versatile Mask-RCNN~\cite{he2017mask} with a resnet50 backbone initialized with weights pretrained on the Microsoft COCO dataset~\cite{lin2014microsoft}.
Therefore, the instance segmentation model classifies each piece as either background or belonging to an instance of food.
We trained the networks on a machine equipped with 384\,GB RAM, an Intel(R) Xeon(R) Gold 6254 CPU @ 3.10GHz, and NVIDIA V100 with 32GB of memory.
Training for each model (initialized randomly for each food) took roughly 24 hours.
Our experiments for inference with trained models were performed on a machine equipped with 31.3\,GB RAM, an Intel Core i7-7700K CPU, and GeForce GTX 1080 Ti. Inference with the trained model during experiments takes about 0.3 sec.

\subsection{Robot Setup}
\label{sec:Robot Setup}
Our robotic system, shown in Fig.~\ref{fig:setup}, consists of a Sawyer 7-DOF robotic arm equipped with a self-designed two-fingered parallel gripper driven by a servomotor (Dynamixel XM430-W350-R).
We use two custom designed end-effectors whose general shape was modelled after tongs commonly used to serve food.
As seen in Fig.~\ref{fig:setup} we call the two tongs-inspired end effectors the ``Fixed Finger'' and a custom designed passive retraction mechanism that allows the fingers to retract when it meets resistance the ``Adaptive Finger''.
The shape of the fixed finger's contact area is the same as the adaptive finger, and both are made of ABS.
Furthermore, we use an Ensenso N35 stereo camera in combination with an IDS uEye RGB camera to overlook the workspace of the robot arm and use them to retrieve registered point clouds of the scene.
We use two industry-standard food trays ($424mm{\times}308mm{\times}160mm$) and designate one to pick from and another to place in.
The Sawyer, gripper and RGB-D sensor are connected to a PC running Ubuntu 16.04 with ROS Kinetic.
We experiment with 7 foods; fried chicken, broccoli, meatball, sausage, mushroom, gyoza (steamed meat dumplings) and taro (colocasia) (Fig.~\ref{fig:simulation_3dmodel}).

\begin{table}[t]
    \vspace{3.0mm}
    \caption{Label agreement between labels obtained on the same image from 5 human annotators (H1-H5) and the trained model. Each row corresponds to taking the labels of that person as ground truth and calculating the Average Precision (IoU=0.5:0.95) with the others in the column.}
    \label{tab:Label Agreement}
    \centering
    \begingroup
    \scalefont{0.80}
        \begin{tabular}{lcccccc}
                                     & \multicolumn{1}{l}{H1} & \multicolumn{1}{l}{H2} & \multicolumn{1}{l}{H3} & \multicolumn{1}{l}{H4} & \multicolumn{1}{l}{H5}  & \multicolumn{1}{l}{Model}\\ \cline{2-7} 
\multicolumn{1}{l|}{H1} &
\multicolumn{1}{c|}{1.0} & \multicolumn{1}{c|}{0.13} & \multicolumn{1}{c|}{\textbf{0.09}} & \multicolumn{1}{c|}{0.08} & \multicolumn{1}{c|}{\textbf{0.04}} & \multicolumn{1}{c|}{\textbf{0.19}}  \\  \cline{2-7}
\multicolumn{1}{l|}{H2} &
\multicolumn{1}{c|}{0.12} & \multicolumn{1}{c|}{1.0} & \multicolumn{1}{c|}{0.04} & \multicolumn{1}{c|}{0.07} & \multicolumn{1}{c|}{0.01} & \multicolumn{1}{c|}{0.11}  \\  \cline{2-7}
\multicolumn{1}{l|}{H3} &
\multicolumn{1}{c|}{0.08} & \multicolumn{1}{c|}{0.05} & \multicolumn{1}{c|}{1.0} & \multicolumn{1}{c|}{0.03} & \multicolumn{1}{c|}{\textbf{0.04}} & \multicolumn{1}{c|}{0.04}  \\  \cline{2-7}
\multicolumn{1}{l|}{H4} &
\multicolumn{1}{c|}{0.07} & \multicolumn{1}{c|}{0.06} & \multicolumn{1}{c|}{0.04} & \multicolumn{1}{c|}{1.0} & \multicolumn{1}{c|}{0.02} & \multicolumn{1}{c|}{0.11}  \\  \cline{2-7}
\multicolumn{1}{l|}{H5} &
\multicolumn{1}{c|}{0.11} & \multicolumn{1}{c|}{0.08} & \multicolumn{1}{c|}{0.06} & \multicolumn{1}{c|}{0.04} & \multicolumn{1}{c|}{1.0} & \multicolumn{1}{c|}{0.06}  \\  \cline{2-7}
\multicolumn{1}{l|}{Model} &
\multicolumn{1}{c|}{\textbf{0.24}} & \multicolumn{1}{c|}{\textbf{0.14}} & \multicolumn{1}{c|}{0.06} & \multicolumn{1}{c|}{\textbf{0.16}} & \multicolumn{1}{c|}{0.03} & \multicolumn{1}{c|}{1.0}  \\  \cline{2-7} 
\end{tabular}
    \endgroup
\end{table}

\section{Experiment Results}
\label{sec:results}
The goal of the experiments is to motivate the use of synthetic data to train the instance segmentation model, verify and compare the performance of the grasping pipeline, as well as understand the effects of the grasp filtering and choice of gripper on attempting to grasp various kinds of foods with a variety of physical properties. 
For our experiments, we choose a variety of real foods that spans a wide range of properties of softness (mushroom $\leftrightarrow$ taro), fragility (gyoza $\leftrightarrow$ taro), regularity in shapes and size (sausage are nearly identical, while pieces of fried chicken aren't ), symmetry (meatballs are symmetric along all planes while broccoli isn't), surface friction (broccoli $\leftrightarrow$ sausage). The list of foods we experiment on includes fried chicken, broccoli, mushroom, meatball, taro, sausage, and gyoza (Fig.~\ref{fig:simulation_3dmodel}).
\begin{figure}[t]
	\centering
	\includegraphics[width=0.9\columnwidth]{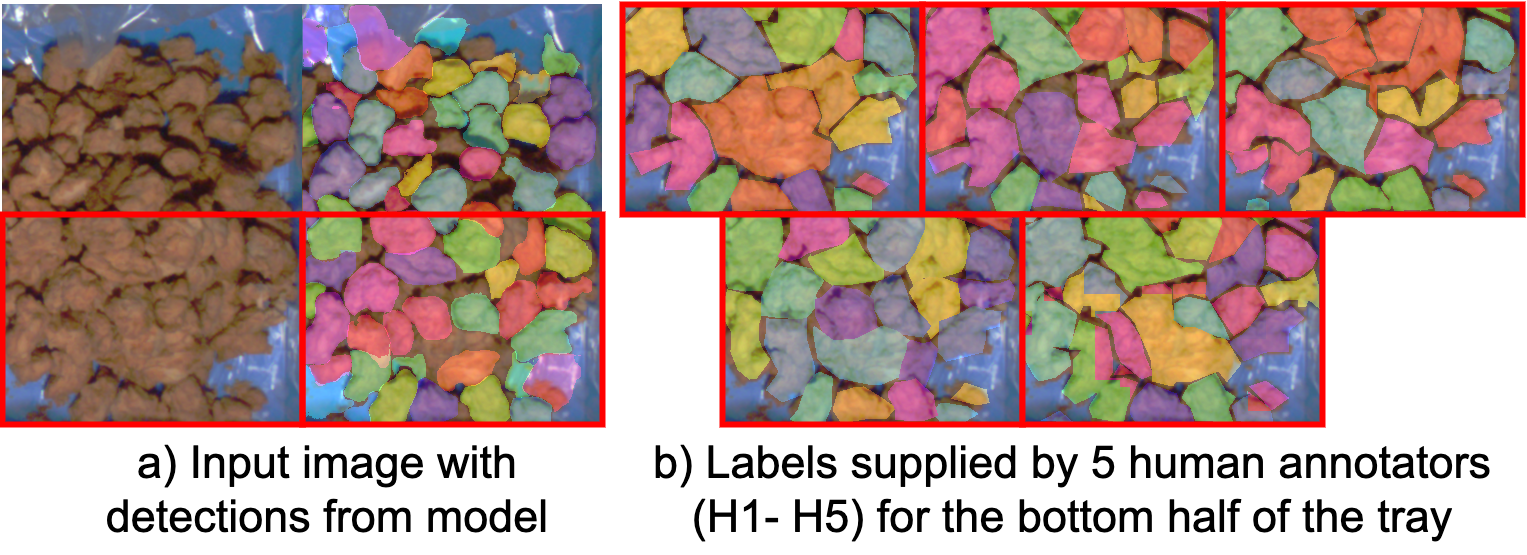}
	\caption{A sample input image with a) predictions from the model and b) labels supplied by the 5 human annotators for the same image.
	}
	\label{fig:disagreement}
	\vspace{-2mm}
\end{figure}

\begin{figure*}[t]
	\centering
	\includegraphics[width=1.6\columnwidth]{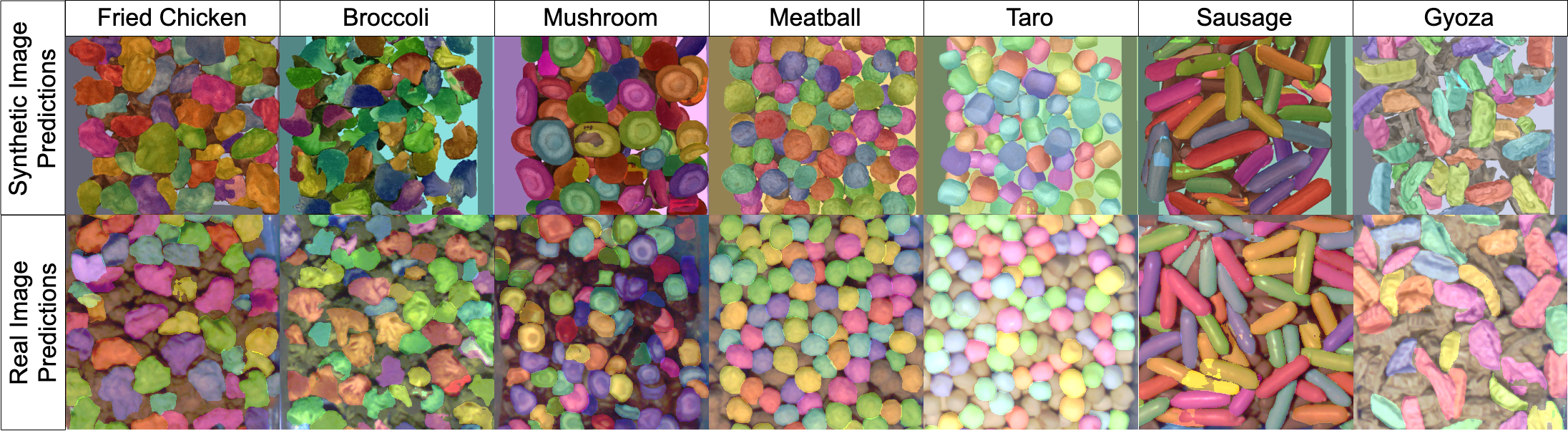}
	\caption{Detections of the instance segmentation model trained with synthetic images for each food and directly applied to real images.}
	\label{fig:allfoods_tray_detections}
	\vspace{-3mm}
\end{figure*}

\subsection{Evaluation of Human Annotation}
\label{sec: results; Annotation Comparison}
\red{In this section, we motivate the use of synthetic data over human annotations by describing difficulties in collection and their propensity for errors.}
Due to limitations in the image quality, a high degree of clutter and the visual similarity of each piece, it is sometimes difficult to identify the boundaries between the pieces. 
This would, understandably, lead to noise in the labeling process, which may confuse the model. 
To verify this hypothesis, we ask 5 unique individuals to label the same image using Amazon Mechanical Turk (AMT) \red{and their provided region annotation web tool}. 
By treating each person's labels as ground truth and comparing them against the others, we can measure the agreement between them as the Precision (IoU between 0.5 to 0.95) of segmentation labels provided. Precision measures the percentage of masks output by someone that overlaps at least 50\% with the ground truth.
As seen in Table~\ref{tab:Label Agreement}, precision values between human annotators H1-H5 tend to be quite low ($<0.1$).
We take this as an indication of noise, or disagreement between individual annotators. 
By visualizing the annotations of the bottom half of the tray in Fig.~\ref{fig:disagreement}, we see a marked difference in how individuals handled areas where boundaries between the pieces are unclear.
We posit that this disagreement may hinder the learning process further indicating a need for the perfect labels we can obtain with the use of synthetic data.

When manual annotations were sourced from AMT, we created a small dataset to be labeled. However, the collection of labels proved harder than anticipated due to many factors. A significant number of annotators submitted zero annotations, despite at least 10-20 pieces being visible at all times. Others labeled all the pieces with the same instance despite providing instructions and a sample image- possibly due to forgetting to switch the ID. A smaller fraction provided inconsistent labels where they started labeling them as multiple instances but then stopped and used the same one for a large number of pieces.
Since hundreds to thousands of images are typically needed for deep learning (especially so to overcome label noise), having these images collected, labeled, verified, and cleaned for every new kind of food would require an enormous investment of time, money and labor. 
If the use of synthetic data is a feasible alternative, the cost required to train these models can be greatly reduced.
\subsection{Simulation Training}
\label{sec: results; simulation training}
The purpose of this section is to show that training on synthetic data provides a practical alternative to training on real images, provides results agreeable with human annotators and is accurate enough to use for grasping.

A direct comparison between training with synthetic data and human-annotated images is difficult due to the reasons discussed in the previous section.
To indirectly evaluate whether our model produces reasonable predictions that a human might make, we adopt the label agreement metric used in the previous section to the model's predictions (taken as ground truth) and the 5 human annotators. From the precision values in bottom line of Table~\ref{tab:Label Agreement}, we see that the model outputs predictions that 3 out of 5 humans most agree with, and the other 2 with reasonable agreement. 

We find that despite being trained purely with synthetic data, the model is able to generalize to predictions in the real world quite well as seen in Fig.~\ref{fig:allfoods_tray_detections}.
We find that it is also robust to changes in lighting conditions, scales and angles and is competent enough to depend on for choosing and executing grasps with high accuracy using the parametrization strategy described in Section~\ref{sec:grasp_detection}.
We note that our end goal is not to train an ideal instance segmentation but one that can be used to grasp food well. 
Even if some pieces of food are not detected, as long as we can find a single reasonable target to grasp, it is acceptable. Therefore, by setting a high detection threshold and only retaining predictions that it is confident in, we should be able to achieve good grasp results even if the model is unable to transfer fully to the real-world data, and the results are a little worse.
\begin{table*}[t]
    \centering
    \begingroup
    \scalefont{0.80}
    \caption{Performance comparison using adaptive and fixed fingers, with and without the grasp filtering. 
    We define success as when a single piece of food is moved from the food tray to the place tray. We also report in brackets the grasping success when more than one piece is deposited in the place tray.
    }
    \label{tab:allresults}
    \begin{tabular}{lllllll}
        &  & \multicolumn{5}{c}{\textbf{Success Rate (incl. multiple pieces)}}\\ \cline{3-7}
        &  & \multicolumn{2}{c}{\textbf{Adaptive Finger}} & & \multicolumn{2}{c}{\textbf{Fixed Finger}} \\ \cline{3-4} \cline{6-7}
        & \textbf{Food Softness}  & \textbf{Grasp Filtering} & \textbf{No Grasp Filtering} & & \textbf{Grasp Filtering} & \textbf{No Grasp Filtering} \\ \cline{3-4}\cline{5-7}
    
        Fried Chicken  & Soft    & 100\% (100\%) & 86\% (86\%)    &&   84\%  (90\%)  & 78\%  (84\%)\\
        Broccoli & Soft     & 96\% (98\%)   & 90\% (96\%)    &&   86\%  (94\%)  & 84\%  (94\%)\\
        Mushroom       & Soft    & 84\% (96\%)   & 80\% (98\%)    &&   64\%  (70\%)  & 48\%  (54\%)\\
        Meatball       & Hard    & 98\% (100\%)  & 90\% (96\%)    &&   96\%  (98\%)  & 96\%  (98\%)\\
        Taro           & Hard    & 92\% (98\%)   & 90\% (98\%)    &&   94\% (100\%)  & 94\% (100\%)\\
        Sausage        & Very hard& 88\% (88\%)   & 82\% (84\%)    &&  100\% (100\%)  & 96\%  (98\%)\\
    \end{tabular}
    \endgroup
    \vspace{-5mm}
\end{table*}

\subsection{Comparison of Adaptive Finger \& Fixed Finger}
\label{sec: expt; hardware}
In this section, we study the effects of using the adaptive finger. 
The success rates of 50 grasps with and without grasp filtering for the adaptive finger and fixed finger are shown in Table~\ref{tab:allresults}.
We define success as when each food item is grasped and put into another food tray.
We also show the incidence where multiple pieces are grasped in a single grasp.
Multiple grasps will be discussed in Section~\ref{sec: expt; grasp filtering}.

\begin{figure}[t]
	\centering
	\includegraphics[width=0.7\columnwidth]{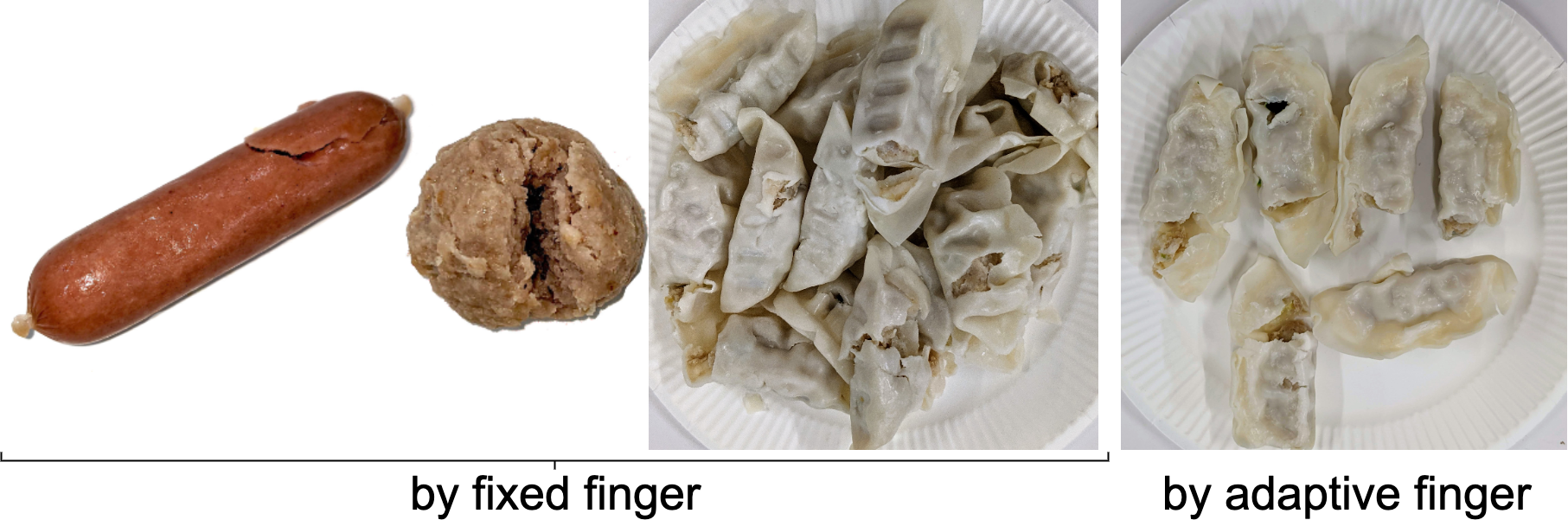}
	\caption{Damage caused by the use of fixed finger on neighbouring foodstuff for sausage, meatball and gyoza (left), by the use of the adaptive finger (right).}
	\label{fig:fooddamage}
\end{figure}

The experimental results in Table~\ref{tab:allresults} show that using the adaptive finger (with and without the grasp filtering) tends to either match or exceed the performance of the fixed finger with the exception of sausage (which we discuss later).
Since the adaptive finger is able to adapt to the surface of the food, it is additionally able to compensate for inaccuracies in instance segmentation, sensing and actuation as visualized in Fig~\ref{fig:soft_schematic} (c).
This improvement in success rate appears most significant in the case of fried chicken $(+16\%)$, broccoli $(+6\%)$ and mushroom $(+20\%)$, which we attribute to the variety in size and surface profiles as well as some compressibility. 
On the other hand, we see a drop with sausage $(-12\%)$, which we attribute to the hardness of the food.
While a fixed finger is able to forcefully push neighbouring or blocking pieces away, the adaptive finger can retract prematurely and simply slip over the target piece due to its smooth, taut skin and may be more suited to be grasped with fixed fingers.
However, we note that this forced grasping may come at the cost of increase in damage done to food via piercing by downward motion or scratching the surface of the food during gripper closure(See ``by fixed finger'' Fig.~\ref{fig:fooddamage}).

While it is difficult to come up with a generic subjective evaluation criterion of damage caused by the grasping process on each foodstuff, we instead choose to measure it specifically for a very delicate food- gyoza, which are meat-filled dumplings.
We do this by counting the number of pieces in both trays with fully pierced skins and whose stuffing is visible after the 50 grasp attempts made in Table~\ref{tab:allresults} for each experiment.
We note these numbers for each method evaluated in Table~\ref{tab:gyoza damage count}.
We see the stark contrast in the number of damaged pieces when the adaptive finger is used instead of the fixed one.
We also note that not just the number but also the amount of damage inflicted on the gyoza is also more significant when a fixed finger is used, with some pieces being practically ripped apart, as seen in Fig.~\ref{fig:fooddamage}. \red{We see that even with the use of adaptive finger, some pieces do get damaged due to a mixture of gripper closure and the spring potentially being too strong for something as delicate as gyoza and could be adjusted to further reduce it.}
The use of the adaptive finger presents a trade-off between performance and damage for some more rigid foodstuff. We believe it is a good default choice based on the foods tested.

\begin{table}[t]
    \caption{Performance comparison and number of pieces damaged by 50 attempts of grasping gyoza.}
    \label{tab:gyoza damage count}
    \centering
    \begingroup
    \scalefont{0.80}
        \begin{tabular}{lp{1.8cm}c}
            \cline{1-3}
            \textbf{Method} & \textbf{Success Rate  (incl. multiple grasping)} & \textbf{\# Damaged Pieces} \\ \cline{1-3}
            Adaptive Finger with Grasp Filtering  & 74\% (90\%)  & 6    \\
            Adaptive Finger without Grasp Filtering  &    56\% (84\%)  &   9  \\
            Fixed Finger with Grasp Filtering    & 76\%  (92\%)  & 14  \\
            Fixed Finger without Grasp Filtering  &   66\%  (94\%)     & 18  \\
        \end{tabular}
    \endgroup
\end{table}

\subsection{Grasp Filtering}
\label{sec: expt; grasp filtering}
In this section, we aim to study the effect of using the grasp filtering.
The success rates of 50 grasps with and without grasp filtering for the adaptive finger and fixed finger are shown in Table~\ref{tab:allresults} as explained in section~\ref{sec: expt; hardware}.

From the results in Table~\ref{tab:allresults}, we see that the use of grasp filtering tends to increase the success rate for most foods for both the adaptive fingers and the fixed fingers.
The use of grasp filtering shows improved performance most prominently for fried chicken $(+14\%)$ and meatball $(+8\%)$.
The fixed fingers also shows improvement to a smaller extent for fried chicken $(+8\%)$ and mushroom $(+16\%)$. 
Since the grasp filtering criterion picks grasps where the fingers are less likely to make contact with neighbouring pieces, we see that it is also able to reduce the incidence of grasping multiple pieces of food with a single grasp. This can be seen from looking at the differences in success rates when multiple pieces are included, which are reported in brackets in Table~\ref{tab:allresults}. This reduction in the percentage of multiple grasps is observed in grasping of foods which are small such as mushroom ($18\% \rightarrow 12\%$) or meatball ($6\% \rightarrow 2\%$) or easily entangled like broccoli ($6\% \rightarrow 2\%$).

The grasp filtering criterion not only improves grasp performance, but in filtering out candidates where collision between the fingers and neighbouring pieces is likely, the amount of damage caused is also reduced during the grasp process.
This is demonstrated in the gyoza experiments in Table~\ref{tab:gyoza damage count}.
We see that using grasp filtering decreases the number of pieces damaged for both the adaptive finger $(9\rightarrow6)$ and the fixed finger $(18\rightarrow14)$.
\section{Conclusion}
\label{sec:conclusion}
In this paper, we propose a method to grasp a piece of target food from a cluttered tray while minimizing the damage to it and its neighbors. Our proposed method consists of 3 components- 1) Generation of synthetic images from 3d models of real food, which we use to train an instance segmentation model and use sim2real methods to aid model transfer to identify real instances of food in the tray. 2) Creation of a novel adaptive mechanism that allows the gripper's fingers to passively retract and avoid damaging food when inserted into the tray. 3) A grasp filtering strategy to evaluate candidate grasps and discard ones that would result in a collision between the gripper and the food and damage neighboring pieces. We confirm the veracity of the first in grasping diverse foods and verify that the use of the adaptive mechanism and the grasp filtering leads to a high success rate of 84\% and above and significantly decreases the amount of damage inflicted on food during the grasping process.

\clearpage
\bibliographystyle{IEEEtran} 
\bibliography{IEEEabrv,bibliography}
\end{document}